%% file: root.tex
\documentclass[letterpaper, 10 pt, conference]{ieeeconf}
\IEEEoverridecommandlockouts
\input{preamble.tex}

\title{\LARGE \bf
Vectorizing Projection in Manifold-Constrained Motion Planning for Real-Time Whole-Body Control
}

\newif\ifanonymous
\anonymousfalse

\ifanonymous
  \author{Anonymous Author(s)}
\else
\author{
Shrutheesh R. Iyer, I-Chia Chang, Andrew Z. Liu, Yan Gu, and Zachary Kingston%
\thanks{SRI, AL, and ZK are with Department of Computer Science, Purdue University, {\tt \{iyer270, liu3447, zkingston\}@purdue.edu}. 
ICC and YG are with School of Mechanical Engineering, Purdue University, {\tt \{chang970,yangu\}@purdue.edu}}%
\\
}
\fi

\begin{document}
\maketitle

\begin{abstract}
Many robot planning tasks require satisfaction of one or more constraints throughout the entire trajectory.
For geometric constraints, manifold-constrained motion planning algorithms are capable of planning collision-free path between start and goal configurations on the constraint submanifolds specified by task.
Current state-of-the-art methods can take tens of seconds to solve these tasks for complex systems such as humanoid robots, making real-world use impractical, especially in dynamic settings.
Inspired by recent advances in hardware accelerated motion planning, we present a CPU SIMD-accelerated manifold-constrained motion planner that revisits projection-based constraint satisfaction through the lens of parallelization.
By transforming relevant components into parallelizable structures, we use SIMD parallelism to plan constraint satisfying solutions.
Our approach achieves up to 100--1000x speed-ups over the state-of-the-art, making real-time constrained motion planning feasible for the first time.
We demonstrate our planner on a real humanoid robot and show real-time whole-body quasi-static plan generation. Our work is available at \url{https://commalab.org/papers/mcvamp/}.
\end{abstract}

\section{Introduction}

Constrained motion planning problems arise commonly in many robotic tasks.
Consider the problem of transporting a cup of water; the robot must find a trajectory in which the mug remains upright throughout to prevent spillage.
Or, consider a humanoid robot transporting a large object.
The robot must maintain balance at all times and keep both arms attached to the box to preserve feasible motion.
Many of these constraints can be represented as implicit functions, which give rise to a submanifold of valid configurations in the robot's configuration space.
Manifold-constrained motion planning is thus the problem of finding feasible, collision-free trajectories that satisfy a constraint imposed by a desired task from a start configuration to a goal region.

\begin{figure}
    \vspace{0.75em}
    \centering
    \includegraphics[width=0.99\columnwidth]{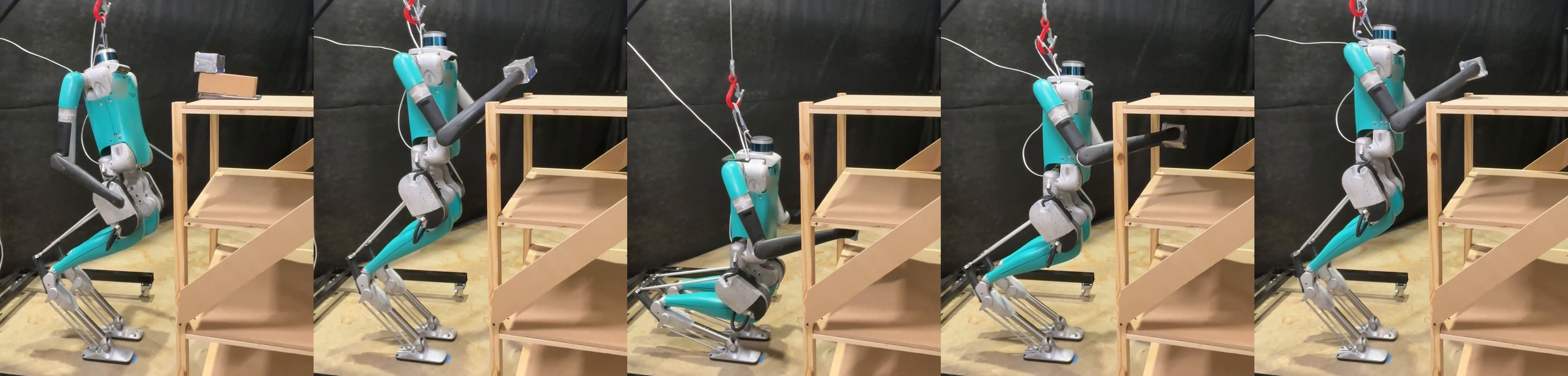}
    \caption{Real-time quasi-static whole-body planning with our vectorized planning approach on the Digit humanoid.
    In this task, the robot must transport a box between the three levels of the shelf, which requires whole body coordination, obstacle avoidance, and constraint satisfaction. Our planner finds feasible motion from start to goal in this scenario in less than 12 milliseconds on average, practical for real-time planning.
    }
    \label{fig:humanoid_transport}
    \vspace{-1em}
\end{figure}

Despite their prevalence, manifold-constrained problems remain difficult to solve in real-time.
The primary bottleneck lies in the ability to find constraint satisfying configurations quickly, as they typically are not explicitly defined.
This problem is also exacerbated by the dimensionality of the system, the number of constraints that must be simultaneously satisfied, and the number of obstacles in the scene.
For example, consider the Digit robot (shown in~\cref{fig:humanoid_transport}), a 30 degree-of-freedom (DoF) humanoid with a 6-DoF floating base. 
In this quasi-static whole-body coordination problem, the robot must maintain balance, have both feet on the ground at all times, and preserve the closed-chain constraint to keep both arms holding the box during transport. 
Consequently, most constrained motion planning works take anywhere in the order of hundreds of milliseconds to tens of seconds to solve, which limits real-time planning and execution.

Recent advances in accelerating motion planning have made real-time motion planning feasible; many approaches have used CPU and GPU acceleration to bring planning time to the millisecond time scale~\cite{sundaralingam2023curobo1,du2025gato,chen2025differentiable}).
Most relevant to our approach is the Vector-Accelerated Motion Planning (VAMP) library~\cite{vamp}, which exploits CPU ``Single Instruction, Multiple Data'' (SIMD) parallelism and data parallelism to achieve multiple orders-of-magnitude speed-ups by parallelizing forward kinematics and collision checking in motion validation, bringing planning times for unconstrained geometric planning problems into the realm of microseconds.
However, while promising, these approaches either do not consider task constraints nor scale to very high-DoF systems.

To this end, we revisit the problem of samping-based manifold-constrained motion planning through the lens of vector acceleration. 
The primary bottleneck---beyond collision checking---is the manifold-constrained extension step, i.e., tracing a geodesic between configurations on a manifold.
We tackle this with \emph{fine-grained parallelization} of the projection of configurations in a geodesic onto the manifold.
Altogether, we present the \textbf{Manifold-Constrained Vector Accelerated Motion Planner (\mcvamp)} which is capable of planning on the order of microseconds to milliseconds for a wide range of manifold-constrained systems including whole-body manipulation.
We believe this is a categorical shift in capabilities of planning for constrained high-DoF systems, as planning in real-time enables reactive global planning which could be used both for high-level task planning as well as low-level control.

Thus, our contribution is as follows
\begin{enumerate}
    \item We present a single-core CPU-only SIMD-accelerated sampling-based manifold-constrained motion planner that is capable of planning in the order of microseconds to milliseconds.
    \item At the heart of our planner is a SIMD-parallel manifold-constrained extension step, which projects multiple configurations in parallel to satisfy constraints, and then evaluates for collision building upon VAMP's existing primitives.
    \item We demonstrate the ability of our planner to compose multiple constraints and develop a quasi-static whole-body motion planner for a humanoid capable of producing stable, feasible motion plans that a robot with a classical controller can execute without modification. 
\end{enumerate}

We evaluate our planner on a number of challenging problems in simulation and reality with a 7-DoF arm, a 14-DoF bimanual system, and a 28-DoF humanoid.
Our approach outperforms the existing state-of-the-art baselines by orders of magnitude in terms of speed, while obtaining a better success rate and similar path quality.

\section{Preliminaries}
The goal of constrained motion planning is to compute a collision-free trajectory between start and goal configurations while adhering to constraints imposed by the robot or task specifications. 
Formally, let $\mathcal{C} \subset \mathbb{R}^d$ be the robot configuration space, where $d$ typically represents the number of actuated joints and $q \in \mathcal{C}$ is a robot configuration. 
The Euclidean space $\mathbb{R}^d$ is referred to as the ambient space. 
The obstacle-free configuration space is represented by $\mathcal{C}_\text{free} \subset \mathcal{C}$.
Let $\mathcal{F}_i(q): \mathcal{C} \to \mathcal{R}^k$ denote the $i^\text{th}$ constraint imposed on the robot, which is a function of the robot's configuration and $k$ represents the dimensionality of the constraint.
A configuration $q$ satisfies the constraint if $\mathcal{F}_i(q) = \mathbf{0}$.
In practice, an acceptable numerical tolerance is imposed, $\mathcal{F}_i(q) \leq \epsilon$.
Each constraint reduces the dimensionality of $\mathcal{C}$, and defines a $(d - k)$-dimensional submanifold $\mathcal{M}_i$ embedded within $\mathcal{C}$.
We also consider the Jacobian $J_i(q)$ of the constraint function $\mathcal{F}_i(q)$, which is the matrix of first-order partial derivatives with respect to $q$.
While we refer to each constraint function individually, we also consider their composition, $\mathcal{F}(q)$, and corresponding composite manifold $\mathcal{M}$.
Thus, the goal of constrained motion planning is to find a collision-free trajectory $\sigma(t): [0, 1] \rightarrow \mathcal{M}_\text{free}$ from $q_a$ to $q_b$ such that it is both collision free and lies on the manifold, given that $q_a, q_b \in \mathcal{M}_\text{free} (=\mathcal{M} \cap \mathcal{C}_{free})$ .

Inspired by~\citet{berenson2011task}, we model end-effector pose constraints as Task Space Region (TSR) constraints.
Let $\mathcal{T} \in SE(3)$ denote a pose transform, where $\mathcal{T}_b^a$ transforms from frame $a$ to $b$, and is composed of a translational and rotational component $[t_b^a, R_b^a]$.
A TSR is composed of two transforms $\mathcal{T}^0_w$, $\mathcal{T}^w_e$, and bounds $\mathcal{B}^w$
\begin{itemize}
    \item $\mathcal{T}^0_w$ transforms global frame to the TSR frame $w$ (i.e., the object frame),
    \item $\mathcal{T}^w_e$ transforms object frame to the end-effector frame,
    \item $\mathcal{B}^w$ represents the bounds for the rotational and transformation errors. This specifies which components of the $SE(3)$ pose of the end-effector are constrained.
\end{itemize}
For example, for a grasped mug to be held upright, $\mathcal{T}^0_w$ is the mug pose, $\mathcal{T}^w_e$ represents the grasp pose, and $\mathcal{B}^w$ specifies that the mug is free to translate and rotate about the $z$-axis, but is tightly constrained along the $x$- and $y$-axes.

Using this formulation, for a given configuration $q$, the constraint function $\mathcal{F}(q)$ is expressed as the distance to the constraint manifold, given by:
\begin{equation}
\label{eqn:tsr}
T_{s'}^w = (T_w^0)^{-1} T_s^0 (T_e^w)^{-1}, \qquad
d^w = \begin{bmatrix} 
t_{s'}^w \\ 
\operatorname{log}(R_{s'}^w)^\vee 
\end{bmatrix} - B^w
\end{equation}

where $T_s^0 = FK(q)$ is the end-effector pose of $q$ obtained by forward kinematics, $T_{s'}^w$ denotes the pose error in $SE(3)$, $d^w \in R^6$, and  $\operatorname{log}(R_{s'}^w)^\vee $ represents the logarithmic map of the $SO(3)$ rotational element of the standard Lie algebra.

\section{Related Work}

Manifold-constrained motion planning has been studied extensively.
Sampling-based planners compute plans by sampling from $\mathcal{C}_\text{free}$ without explicitly computing the space, including multi-query planners like PRM~\cite{kavraki1996prm} and single-query planners like RRT~\cite{kuffner2000rrtc}.
These methods extend to constrained planning by sampling constraint-satisfying configurations, which fall into two major categories of approach.
Projection-based methods use the constraint Jacobian to find satisfying configurations, e.g., CBiRRT2~\cite{berenson2011task} uses Jacobian pseudoinverse projection within bidirectional RRT. 
Continuation methods construct approximations of the manifold using the Jacobian to construct tangent spaces, such as in AtlasRRT~\cite{jaillet2013atlasrrt}.
We refer the reader to~\citet{kingston2018ar} for a full survey. 
Despite progress, efficient manifold sampling remains a bottleneck for real-time planning.

Learning-based methods show promise for acceleration, e.g. CompNetX~\cite{qureshi2020neural} and C-NTFields~\cite{ni2024physics} learn to sample on task-specified manifolds or learn the manifold structure. 
However, these require large amounts of training data and cannot easily adapt to new constraints.
Another approach samples directly from manifolds using inverse kinematics (IK)~\cite{wang2019inverse}, though IK computation often bottlenecks performance.
\citet{cohn2024constrained} reformulate bimanual constraints using an analytic IK formulation which reparameterizes the constraint, enabling planning through the Graph of Convex Sets (GCS) method~\cite{marcucci2023motion}. 
However, this restricts solutions to a single C-bundle and requires extensive pre-computation.

\subsection{Parallelized Motion Planning}
There are many approaches to parallelizing motion planning; multiple aspects of planning are trivially parallelizable such as running many searches~\cite{cforest} or many iterations in parallel~\cite{ichnoprrt,huang2025prrtcgpuparallelrrtconnectfast}.
GPU approaches~\cite{huang2025prrtcgpuparallelrrtconnectfast} merge multiple parallelism levels.
The cuRobo~\cite{sundaralingam2023curobo1} planner uses GPU-parallel multi-seed trajectory optimization, imposing soft penalties for constraint adherence as is typical~\cite{schulman2014motion,ratliff2009chomp}
However, soft formulations do not guarantee strict manifold adherence.

Our approach builds off of VAMP~\cite{vamp}, which performs vectorized collision checking through CPU SIMD, showing tremendous speedup.
Another recent work, cPRRTC~\cite{hu2025cprrtcgpuparallelrrtconnectconstrained}, showed that parallelizing projection can accelerate constrained motion planning.
Our approach demonstrates that VAMP's insight of batching validation across an edge holds also for manifold projection, providing order-of-magnitude speedups over baseline approaches.

\subsection{Whole Body Planning}
Whole-body humanoid planning must satisfy multiple constraints (balance, collision avoidance, task poses) for high-DoF systems (>25 DoF), making these spaces highly non-convex.
For multiple constraint satisfaction,~\citet{sentis2005synthesis} use hierarchical null-space projection, which does not handle conflicting constraints well.
\citet{kaiser2012constellation} introduce the Constellation technique for conflicting constraints, which~\citet{agrawal2025constrained} build upon by proposing constrained nonlinear Kaczmarz for large constraint sets.
We use cyclic projection in our approach to handle many constraints simultaneously.

Specific to humanoids, planners such as CBiRRT~\cite{berenson2011task} have tackled the problem, but are not fast enough for real-time scenarios.
\citet{lembono2021learning} speed up planning by learning feasible configurations of the system, but require large amounts of training data.
IK-based reactive control such as ~\cite{atkeson2015no} are fast but often get stuck in local minima.

\section{Methodology}

In manifold-constrained sampling-based motion planning, local connections between configurations should satisfy constraints.
Typically, these are approximations of a geodesic on the constraint submanifold, and are computationally expensive to generate.
Our primary contribution is vectorizing the generation of constraint satisfying motion by parallelizing the projection step that maps configurations onto the constraint manifold.
The key insight is that multiple configurations along an extension can be projected and checked in parallel using SIMD operations, dramatically reducing the overall cost of constraint satisfaction.
We first present the complete manifold-constrained RRT-Connect algorithm, then detail the vectorized projection mechanism, and finally discuss the choices that went into the design of our planner.

\input{crrtc_algo.tex}
\input{parallelconstrainedextend.tex}

\subsection{Vectorized Projection-Based Sampling-Based Planning}

\begin{figure*}[!t] 
\centering
    \includeinkscape[width=0.99\linewidth]{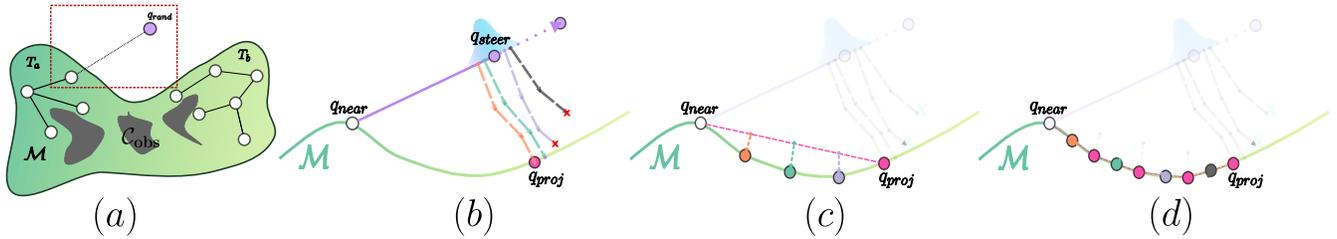}
    \caption{Methodology. (a) RRT-Extend Step: A random configuration in ambient space is sampled. (b) $q_{steer}$ is computed at a fixed distance from $q_{near}$ and samples around $q_{steer}$ are projected onto the manifold until any one succeeds. (c) Interpolated samples along the vector connecting the start and the initial projected point are projected in parallel. (d) Configurations are interpolated between the projected particles and are projected and validated recursively until desired resolution is achieved. \textit{$n=4$ here for illustrative purposes, each represented by a different color.}}
    \label{fig:methodology}
    \vspace{-1em}
\end{figure*}

Our parallel vectorized manifold-constrained RRT-Connect algorithm is presented in \cref{alg:crrtc}.
This is a variant of the RRT-Connect algorithm~\cite{kuffner2000rrtc} where all nodes and edges lie on the manifold $\mathcal{M}_\text{free}$ up to some discretization resolution $\delta$.
In practice, we also use the dynamic-domain~\cite{Jaillet2005} and balancing heuristics~\cite{kuffner2005balanced}, but have elided these from the pseudocode for clarity.

First, a $q_{rand}$ is sampled from the ambient space, and its nearest neighbor $q_{near}$ is found (~\lineref{alg:crrtc}{algline:crrtc:near}).
The \contrib{ParallelConstrainedExtend} method attempts to connect $q_{near}$ to $q_{rand}$ (~\lineref{alg:crrtc}{algline:crrtc:extend}).
If the extension is successful, the planner repeatedly attempts to grow the projected configuration toward the goal tree using the same \contrib{ParallelConstrainedExtend} method with a fixed extension step size.

The key contribution of our work is the vectorized \contrib{ParallelConstrainedExtend} method (\cref{alg:parallelextend}). Following the intuition and empirical results from VAMP~\cite{vamp} we focus on the constrained motion validation step, and parallelize the projection and validation of interpolated configurations.
However, this is not trivial since each point along the extension vector (i)~could be at different distances from the manifold, and (ii)~have to be projected such that there is a continuous path on the manifold between the projected points. 
To deal with these issues, we present a two step vectorized approach. 
Throughout the rest of the discussion, $n$ denotes the number of parallelized operations.

First, $q_{steer}$ (\lineref{alg:parallelextend}{algline:ext:steer}) is computed at a fixed distance away from $q_{near}$ along the vector $q_{nr}$. Then we sample $n$ points around $q_{steer}$ (\lineref{alg:parallelextend}{algline:ext:particles}) on direction vector, and attempt to project all points onto the manifold. We exit if any of them succeed and record the particle that succeeds the earliest, as $q_{proj}$ (\lineref{alg:parallelextend}{algline:ext:proj}). 
Then, $n$ points are linearly interpolated between $q_{near}$ and $q_{proj}$ and projected onto the manifold (\lineref{alg:parallelextend}{algline:ext:interp}) in parallel.
The next batch of $n$ points are recursively interpolated and projected between adjacent projected points until the desired motion validation resolution is achieved. 
Collision checking is performed in parallel after projection, and we exit early if any projected point is in collision.

Sine the number of points to check along the motion vector is typically small (on the order of 8--64), this is perfectly suited to SIMD parallelization, which provides a high throughput without significant overhead, and allows flexible interleaving of sequential and parallel code, something that an iterative optimization method benefits from. 
To support data parallelism necessary for SIMD instructions, all configurations, function evaluations, and Jacobians are stored in a structure-of-arrays layout.

The critical component of the parallel extend method is parallel projection of multiple points onto the manifold. To be able to perform this, a few pieces are needed. 

\noindent\textbf{Tracing Compiler for Constraints}
In a manner similar to VAMP, we generate SIMD parallel code for evaluating constraint functions using a tracing compiler to \emph{trace} the low-level operations needed to compute the \texttt{distanceToConstraint} (\lineref{alg:parallelextend}{algline:ext:distance}).
We use build upon the existing tracing compiler used by VAMP which uses Pinocchio~\cite{carpentier-sii19} and CppAD~\cite{cppad} to generate efficient, branch-free, loop-unrolled code for robot kinematics, which is amenable to SIMD operations.
By using automatic differentiation during the tracing process, we efficiently compute the corresponding Jacobian matrices of the constraint functions.
To enable gradient-based optimization, we implement a differentiable version of $\operatorname{log}(R_{s'}^w)^\vee$ by using a first order Taylor Series approximation of the \texttt{sinc} function at singularities.

\noindent\textbf{Vectorized Levenberg–Marquardt:}
In addition to tracing the constraint functions and Jacobians, we also subsequently trace one step of the Levenberg-Marquardt (LM) algorithm to be able to more efficiently perform the gradient descent in projection.
We exploit the known dimensionality of the robot and the constraint functions by using a second-order LM algorithm.
To achieve this, we trace compile the Jacobian matrix pseudoinversion for each constraint. 
Given the semipositive definite (SPD) formulation of the LM step, we implement a custom Cholesky decomposition method~\cite{dahlquist2003numerical}, which also follows the SIMD principle of branchless control flow, and the fixed dimensionality allows us to perform loop-unrolled Cholesky solve expressions that can be parallelized. 
This subroutine is compiled for each constraint and allows us to perform one LM step in parallel for multiple particles. 
We provide both the inner and outer matrix pseudoinversions, the former is bounded by the dimension of the manifold function while the latter is bounded by the dimension of the robot.

\subsection{Design Choices}

Vectorized projection provides a twofold benefit. First, we can quickly project and evaluate multiple configurations in parallel, which provides naively an $n$-times speed-up compared to the sequential counterpart. More importantly, the parallel step also allows us to invalidate infeasible configurations very quickly, reducing wasted computations, which leads to the empirical 100-times speed-up we see in our experiments. 
By checking future configurations further along the validate step, we can cheaply invalidate the entire motion if any particle fails to project or is in collision. This follows the ethos of sampling-based planning by quickly invalidating infeasible motion to focus search elsewhere.

\noindent\textbf{Early Exits}
The independent parallel nature of projection can cause inconsistencies where the interpolated points project to different parts of the manifold. Validating the connection between these points requires further interpolation, which may become unbounded. 
We avoid this altogether by exiting projection early if any particle has a large descent step, which likely indicates that the particles may be quite far from each other. \lineref{alg:parallelextend}{algline:ext:earlyexit1} perform this early exit at different levels and invalidate the entire motion validation if any particle violates a projection distance threshold.

\noindent\textbf{Two-Stage Projection}
While the two-stage projection may seem unnecessary at first (as prior projection-based planners simply interpolate towards $q_{steer}$), for hard constraints this reduces the likelihood of interpolated points projecting far away from each other during descent, as they are already quite close to the manifold. We provide an ablation study of this in \cref{sec:bimanual}. 
The two-step projection also behaves as an adaptive range parameter for extension step: by sampling points around $q_{steer}$, depending on the difficulty of the constraint, a point that is either closer or farther away from the manifold could be projected the earliest, and overall provides higher chance of successful extension.

Consequently, our tree growth approach is built with the following principles: (i) stay close to the manifold while extending to limit the amount of computationally expensive projection iterations, and (ii) employ particle-based optimization to improve the success of manifold sampling, as even when the initial seeds are proximal in the ambient space the local gradient landscape can lead to divergent behavior.

\begin{figure}
    \centering
    \includeinkscape[width=0.7\linewidth]{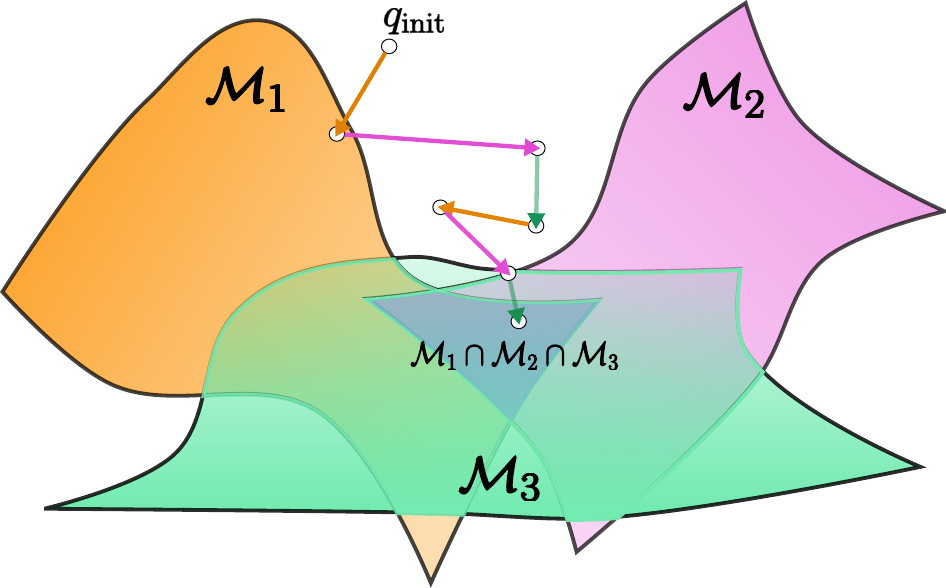}
    \caption{Cyclic Projection to intersection of manifolds. At each iteration, we take one step towards each manifold in a pre-determined order, and the algorithm stops when we reach their intersections}
    \label{fig:intersectionmanifolds}
    \vspace{-1em}
\end{figure}

\subsection{Composition of Constraints}
Some problems require satisfying multiple constraints, e.g., a humanoid carrying an object, as we explore in~\cref{sec:humanoid}.
Thus, we need to be able to support multiple constraints, which translates to projecting onto the intersection of the manifolds, as seen in \cref{fig:intersectionmanifolds}.
Naively, these constraints can be combined into one large system of equations, but this scales poorly, as noted by~\cite{agrawal2025constrained}.
We use a cyclic projection technique~\cite{214546}, described in~\lineref{alg:parallelextend}{algline:cyclic}. During each iteration of optimization, we compute the descent direction of one constraint and takes a step in that direction. 
Constraints are cycled through in a predetermined order, and this step is repeated until we converge onto the intersection of all constraint manifolds. 
Although it is a simple approach, it empirically works well, since its simplicity allows parallelization at no additional cost.

\section{Experimental Evaluation}

We evaluate our planner on a wide suite of problems with different types and complexities of constraints. 
We test it out on 3 different robots: the Franka Emika Panda Arm (7-DoF), a bimanual Kuka IIWA system (14-DoF), and the Digit Robot (36-Dof, including a 6-DoF floating base). 
We primarily evaluate our planner against the projection-based manifold-constrained RRT-Connect implementation from \ompl ~\cite{sucan2012ompl, kingston2019exploring} and \curobo~\cite{sundaralingam2023curobo1} for the Panda arm.
We perform hyperparameter sweeps on all planners on each problem and choose the best performing configuration.
For fair comparison, \ompl uses the collision checking of VAMP; constraint evaluation is done using Pinocchio~\cite{carpentier-sii19}.
For the Bimanual KUKA IIWA arm, we compare against the approach of~\citet{cohn2024constrained}.
All algorithms were tested on an 5.4Ghz Intel i7-13700K CPU with 64GB of RAM and an NVIDIA RTX 4090 GPU with 24GB VRAM (for \curobo).

\subsection{End-Effector Constraint for a Single Arm}

\begin{figure}
    \centering
    \includegraphics[width=0.85\linewidth]{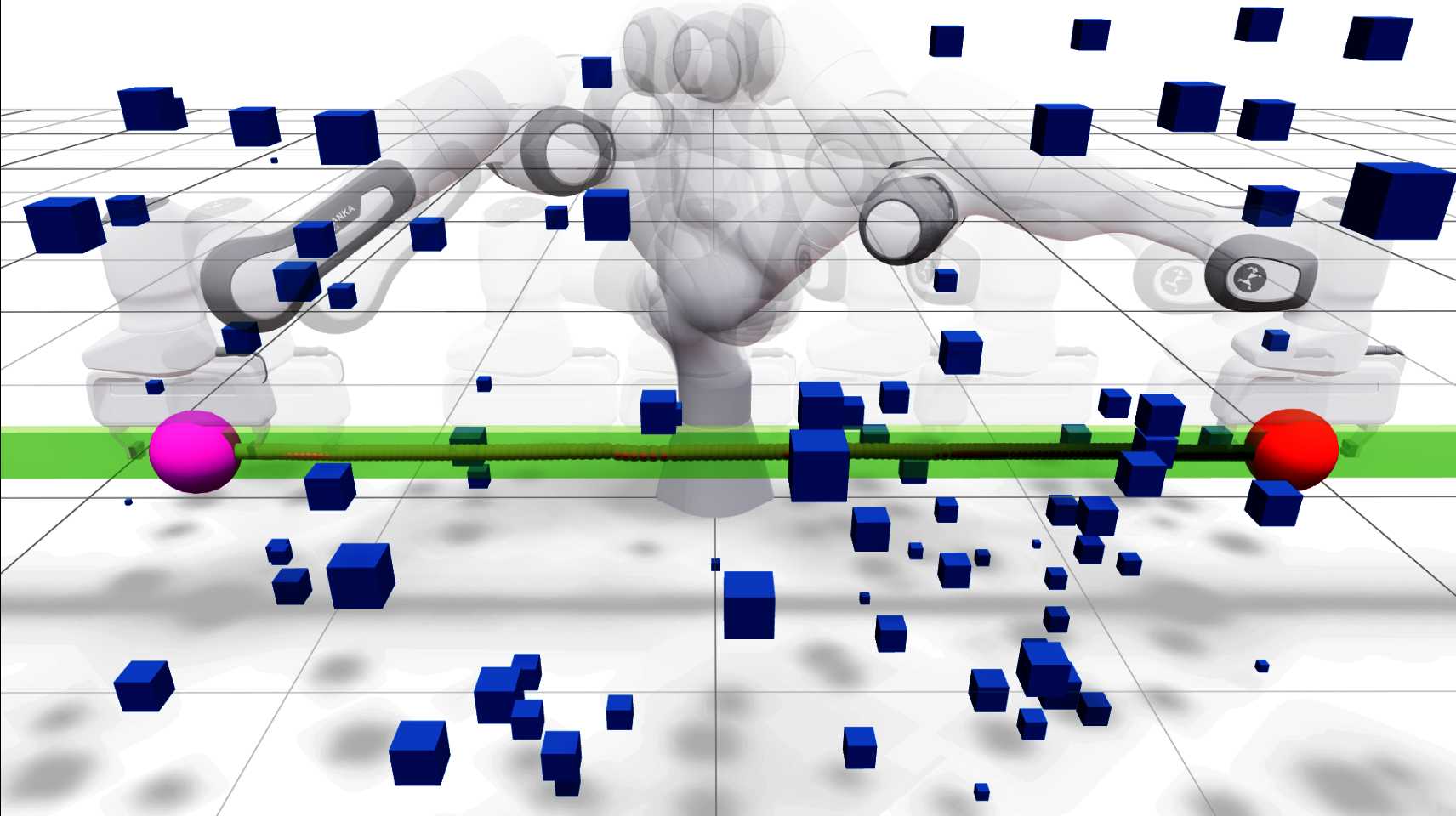}
    \caption{Example of a line problem generated with start goal pairs (spheres) and the solution. Random cuboid obstacles are generated around the robot which have to be avoided.}
    \label{fig:line_plane_prob}
    \vspace{-1em}
\end{figure}

\input{line_plane_figs}

We first evaluate our constrained planner on the 7-DoF Franka Emika Panda arm. 
We test on 4 constraints: keeping the end-effector on a plane (i) without an orientation constraint and (ii) with a fixed orientation, and keeping the end-effector on a line (iii) without and (iv) with an orientation constraint.
An example problem can be seen in \cref{fig:line_plane_prob}.
To generate the problems, we sample from a set of pre-defined lines and planes as constraints and sample valid start and goal configurations on the constraint. 
For each start-goal pair we incrementally add obstacles in the task space. 
Finally, we run our planner with many different hyperparameters and retain the problem even if any succeed.
Note that \curobo's constrained planner does not allow providing joint space configurations goals; we provide the end-effector pose of the goal configuration.
We set a timeout of 10 seconds for \ompl after which the planner reports failure.

\Cref{fig:line_plane_plots} shows the results for each class of constraint problem over a range of obstacle densities.
In general, across all benchmarks we obtain superior performance in terms of planning time, with 100\% success even as the difficulty increases. 
Most plans are under 1ms, which is anywhere from 100--2000$\times$ faster than the \ompl and \curobo baselines. 

\begin{figure}
    \centering
    \begin{subfigure}{\columnwidth}
        \centering
        \includegraphics[width=0.75\linewidth]{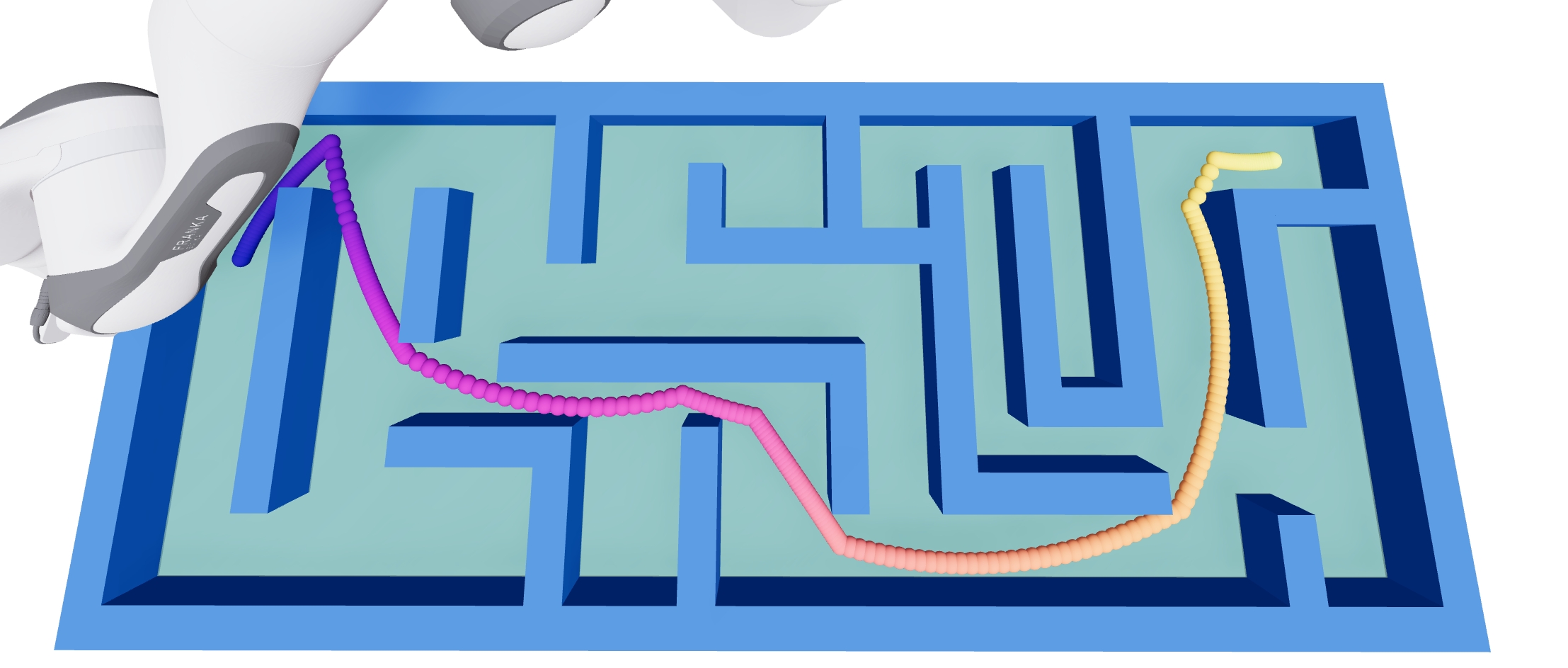}
        \caption{Generated plan to solve the maze}
        \label{fig:maze_plan}
    \end{subfigure}
    \begin{subfigure}{\columnwidth}
        \centering
        \includegraphics[width=0.75\linewidth]{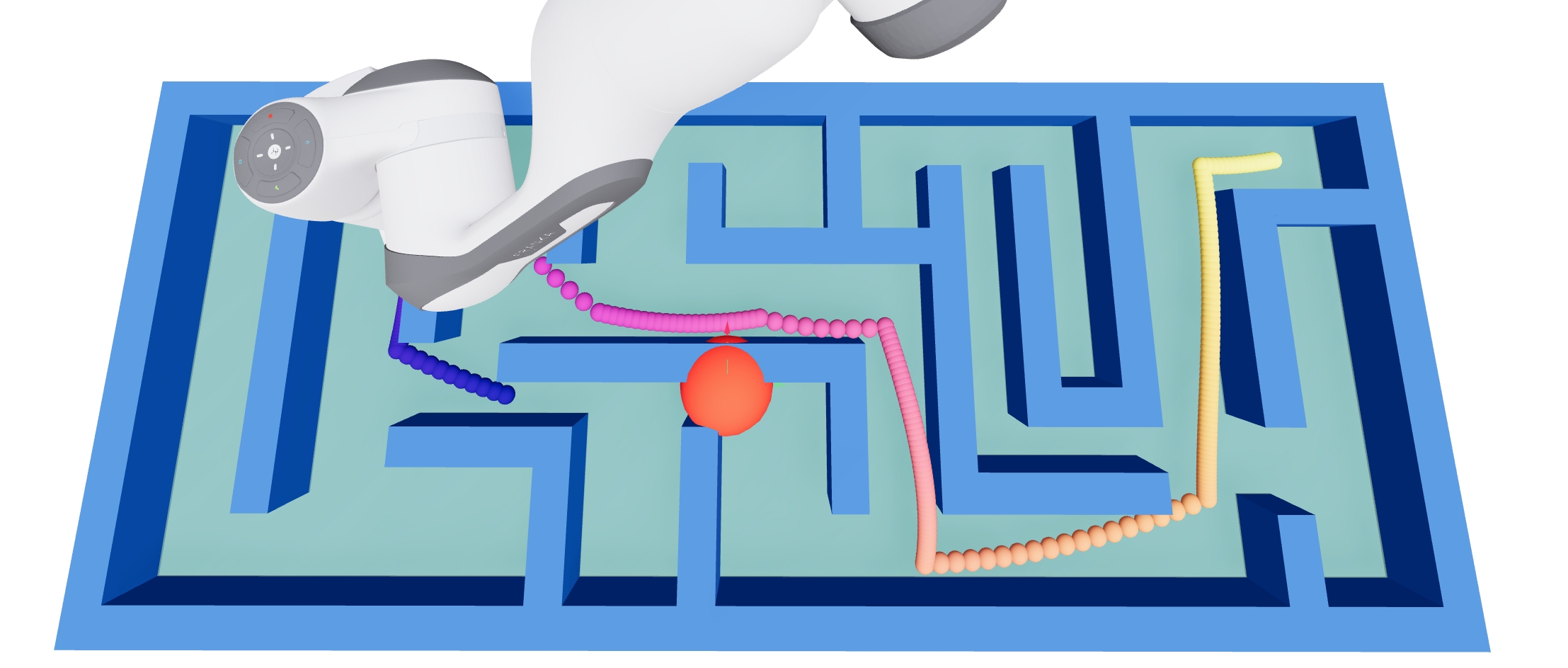}
        \caption{Real-time replanning for dynamic environments}
        \label{fig:maze_replan}
    \end{subfigure}
    \caption{Solving a Maze. Here the tip of the marker is constrained to the floor of the maze. Consequently, the robot has to solve the maze to find a plan from start to goal. Planning is done at 20Hz. (b) when a dynamic obstacle blocks the path (red sphere here), the planner can compute a new path due to the high motion validation and planning throughput}
    \label{fig:maze_problem}
\end{figure}

Next, we evaluate the effectiveness of our planner to navigate complex environments while respecting constraints using a maze (shown in \cref{fig:maze_problem}). 
We initialize 100 random start goal pairs at different points on the maze a minimum distance apart.
We attach a marker to the end effector of the robot, whose tip is constrained to the bottom plane of the maze. 
In addition, we also impose an orientation constraint, such that only the yaw is free (i.e., the marker is free to rotate about its axis). 
We benchmark our algorithm against \ompl. As an ablation study, we replace the Pinocchio-based projection of \ompl with our compiled projection. 
For \ompl, we set the timeout to 100 seconds.

From the results in \cref{tab:maze_table}, our planner shows a remarkable improvement on planning times and success rates, achieving over a 1000$\times$ speed-up in some cases, opening the door to real-time planning for hard manifold constraint problems.
We illustrate this in \cref{fig:maze_replan} where the planner is able to avoid a dynamic obstacle that blocks its current path.

\input{maze.tex}

\subsection{Bimanual Arm Constraint}
\label{sec:bimanual}

To test the scalability of the planner to higher dimensions, we evaluate it on a box transport problem with a bimanual system. 
In this task, the 14-Dof arms are constrained such that the relative transform between the left and the right arm must be fixed throughout the motion, i.e., holding the box.
To enforce this constraint, we formulate it as a TSR constraint:
\begin{equation}
    \begin{aligned}
        T_{err} &= T_{ref}^{-1} (T_{left}^{-1} T_{right}), &
        d_{err} &= \begin{bmatrix} t_{err} \\ \operatorname{log}(R_{err})^\vee \end{bmatrix}
    \end{aligned}
    \label{eqn:bimerr}
\end{equation}

Following~\citet{cohn2024constrained}, we evaluate our planner on the same problem of moving the arms between shelves while maintaining the relative pose and avoiding collisions.
For baselines, we use compare against both their IK-BiRRT planner as well as the IK-GCS approach; these both use a pre-computed analytic representation of the manifold constraint. 

As this problem is more complex and higher-dimensional, we evaluate an ablation our two stage projection approach to prove its value.
We test a single-stage projection, where points are interpolated between $q_{near}$ and $q_{steer}$ (\mcvamp 1-step). \Cref{tab:bimanual} shows the results of the planners. 

Our approach provides 30--60$\times$ speed-up for solving the task compared to IK-BiRRT, and more than a 10$\times$ speed-up compared to IK-GCS. 
This result is interesting as IK-GCS requires pre-processing (about 70 seconds) to compute the graph of convex sets, which we do not count for benchmarking. 
However, IK-GCS produces shorter paths, owing to their optimization-based formulation. 
Our 2-step projection also outperforms the 1-step parallel projection approach, which indicates that it is beneficial as the complexity of the constraints increase.

\input{bimanual.tex}

\subsection{Whole Body Planning}
\label{sec:humanoid}

\begin{figure}
    \centering
    \begin{minipage}{0.3\linewidth}
        \centering
        \includegraphics[width=\linewidth,trim=1em 0em 1.5em 0.5em,clip]{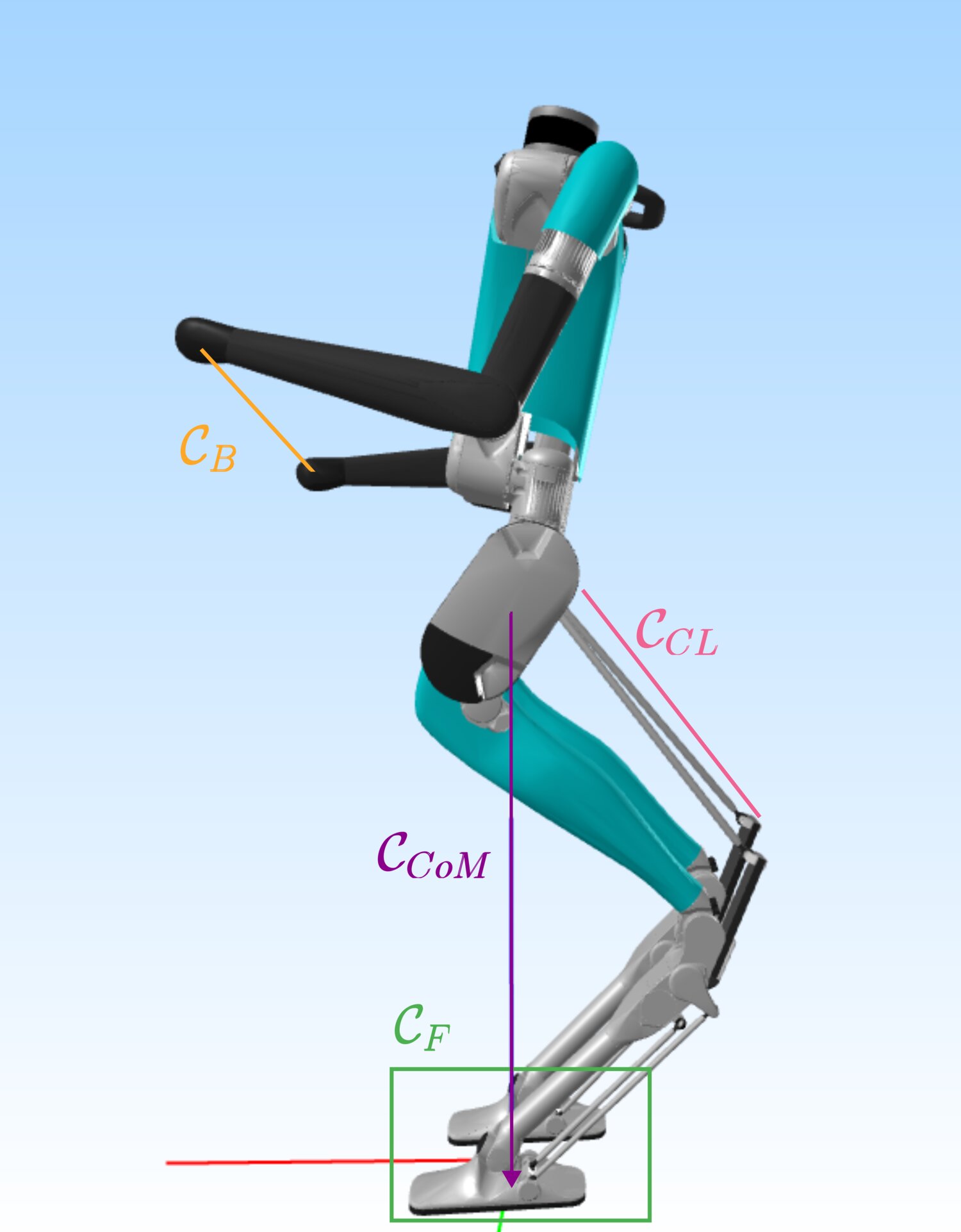}
    \end{minipage}
    \begin{minipage}{0.65\linewidth}
        \centering
        \footnotesize
        \textbf{Constraint Equations} \\
        \scriptsize
        \renewcommand{\arraystretch}{1.2}
        \begin{tabular}{@{}rl@{}}
            \bottomrule
            $\mathcal{C}_B$: & $T_{\text{err}} = T_{\text{ref}}^{-1}(T_L^{-1}T_R)$ \\
             & $d_B = \begin{bmatrix} t_{\text{err}} \\ \log(R_{\text{err}})^\vee \end{bmatrix}$ \\[0.5em]
            \midrule
            $\mathcal{C}_{CL}$: & $d_{CL} = \begin{bmatrix} \|X_{\text{hip}}^l - X_{\text{tar}}^l\| \\ \|X_{\text{hip}}^r - X_{\text{tar}}^r\| \end{bmatrix}$ \\[0.5em]
            \midrule
            $\mathcal{C}_{\text{CoM}}$: & $x_{\text{near}} = \mathrm{proj}(\mathbf{x}_{\text{com}}^{xy}, P)$ \\
             & $d_{\text{COM}} = \mathbf{x}_{\text{com}}^{xy} - x_{\text{near}}$ \\[0.5em]
            \midrule
            $\mathcal{C}_F$: & $d^f = \begin{bmatrix} t_{\text{rel}}^f \\ \log((T_{\text{ref}}^f)^{-1}T_f)^\vee \end{bmatrix}$ \\
             & $d_F = \begin{bmatrix} d^l & d^r \end{bmatrix}^\intercal$ \\
            \bottomrule
        \end{tabular}
    \end{minipage}
    \caption{The different constraints of the Digit robot. (i) $\mathcal{C}_B$, the bimanual fixed pose constraint, (ii) $\mathcal{C}_{CL}$, the closed linkage constraint, (iii) $\mathcal{C}_{COM}$, the center of mass constraint, and (iv) $\mathcal{C}_F$, the fixed feet placement constraint.}
    \label{fig:digit_constraints}
    \vspace{-1em}
\end{figure}

In this scenario, we scale beyond beyond constraints on just the end-effector of a fixed-based manipulator and evaluate the ability to tackle composition of constraints. 
We implement our planner for the Digit robot to perform quasi-static whole body planning to transport a box. 
The Digit robot has 30 joints and 6-DoF floating base.
Of these 30 joints, 4 are high-stiffness springs, while another 6 are passive and constrained by closed-loop linkages.
By assuming rigidity of the 4 spring joints and relying on the low-level tracking controller to handle the 4 closed-loop linkages associated with the ankles, we model it as a 28-Dof system, composed of 22 joints and a 6-Dof floating base, where 2 joints are passive and constrained by 2 closed-loop linkages.
To generate a feasible motion plan for the Digit, the motion must respect the following four constraints (equations detailed in~\cref{fig:digit_constraints}):
\begin{itemize}
    \item \textbf{Stability Constraint:} To maintain balance, the $xy$-projection of the center of mass (CoM), $\mathbf{x}_{com}^{xy}$, must remain within the convex hull of the support polygon $P$.
    The error is to the nearest point in the hull.

    \item \textbf{Feet Constraint (TSR):} For fixed feet placement, we stack the 6-DoF pose errors for both feet ($f \in \{l, r\}$) into a 12-element vector.

    \item \textbf{Closed Link Constraint:} The closed loop linkages between the hip and the tarsus joint are enforced as fixed lengths between the two links.

    \item \textbf{Bimanual Constraint:} The relative pose between the hands is fixed, same as in~\cref{sec:bimanual}.
\end{itemize}

Satisfying the first three constraints is essential for the Digit, as they are required for stability and real-world operation.
To control the robot to track the planned joint trajectory, a passivity-based controller~\cite{gong2022zero} is used to generate feedforward motor torque commands that compensate for the ground reaction forces, and uses proportional-derivative feedback to correct tracking error.

We evaluate our planner in two real-world experiments.
In additional to planning time, we also verify successful planning by reporting the tracking errors of relevant joints.
A small tracking error indicates that the motion plan is more dynamically feasible, as the controller does not need to adjust motion to respect the stability, feet, and closed-link constraints.
For these tasks, the planner was run on a 4.8GHz Intel Core Ultra 7 258V device with 32GB of RAM.

\subsubsection{Box Transport}

For this task, the robot is required to transport the box between three levels of the shelf, shown in~\cref{fig:humanoid_transport}.
Each task is repeated 3 times; \cref{tab:humanoid_shelf_planning} shows the planning times for each task.
For motion segment, we show the distribution of the tracking error of the highest offending joints (the knee and the tarsus joints).
Errors range less than 10 degrees on average, with a max error across all joints of 12 degrees at a single point for the left tarsus joint.

\input{humanoid_transport}

\subsubsection{Dynamic Obstacle Avoidance}

\begin{figure}[t]
    \centering
    \begin{subfigure}{\columnwidth}
        \centering
        \includegraphics[width=0.95\columnwidth]{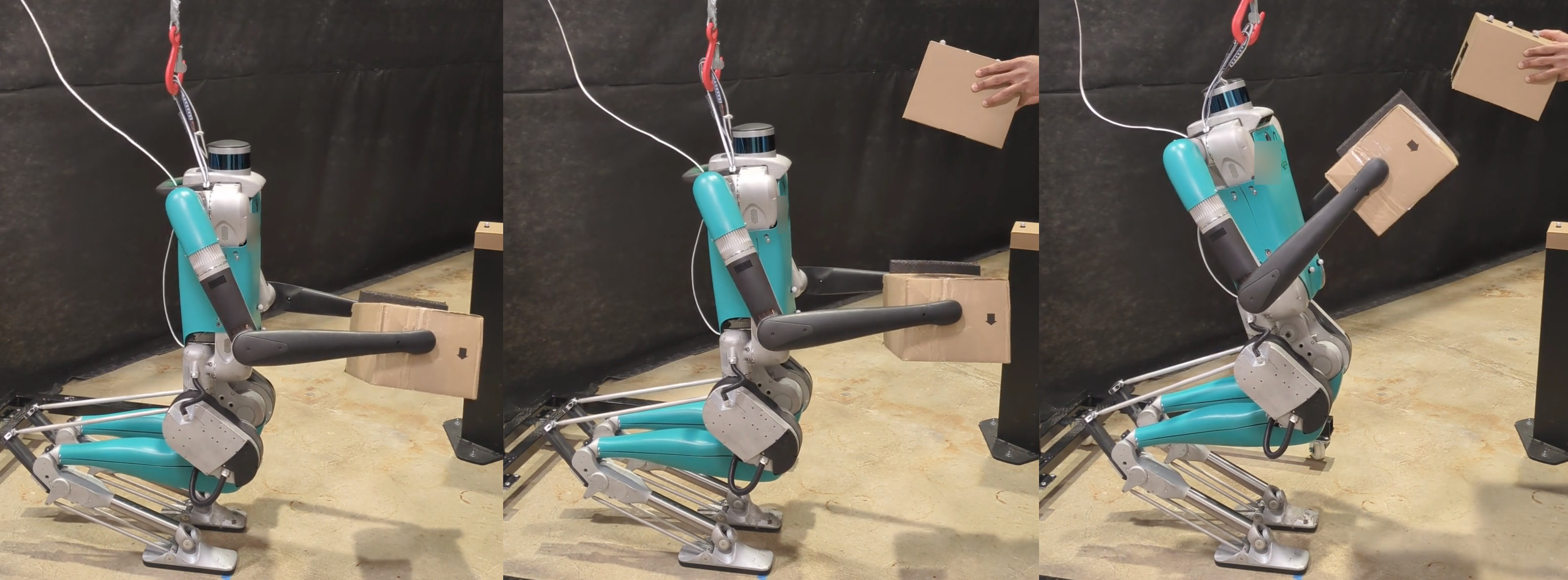}
        \caption{Whole-body trajectory execution in dynamic environment. During the upward motion, an obstacle is detected, the robot plans a feasible trajectory around the obstacle. }
        \label{fig:obstacle_dodge}
    \end{subfigure}
    
    \vspace{1em}
    
    \begin{subfigure}{\columnwidth}
        \centering
        \includegraphics[width=0.9\columnwidth]{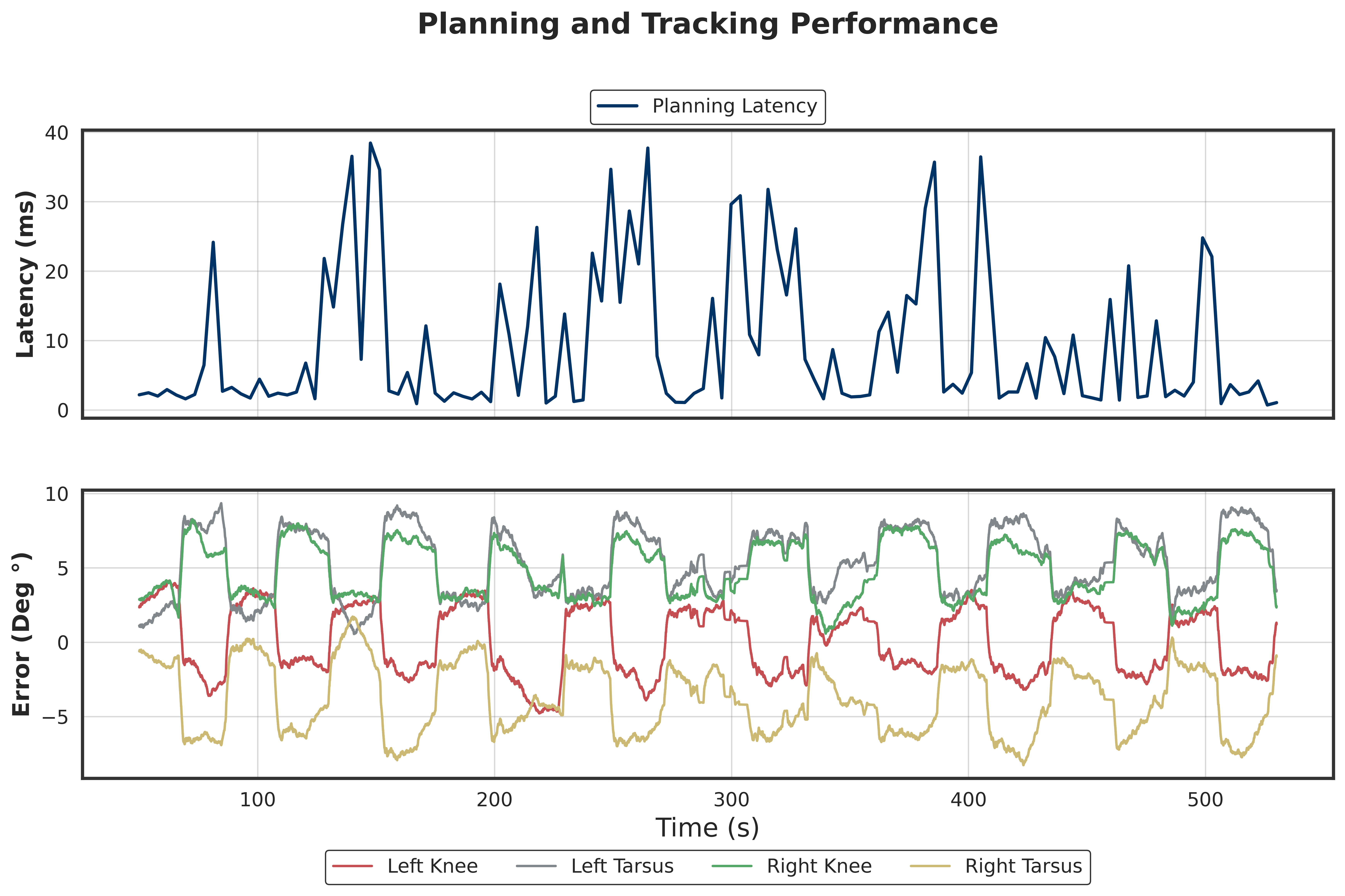}
        \caption{Tracking error and planning times for the repeated start-goal-start motions for the bimanual}
        \label{fig:planning_times_obs_dodge}
    \end{subfigure}
    \caption{Experimental results for dynamic environment obstacle avoidance.}
    \label{fig:full_experiment}
    \vspace{-1em}
\end{figure}

Here, the humanoid robot is tasked to repeatedly plan and execute a box transport task between fixed locations. 
A dynamic obstacle is introduced, and we test the ability of the robot to react and avoid the obstacles.
\Cref{fig:obstacle_dodge} illustrates the motion of the robot, while \cref{fig:planning_times_obs_dodge} shows the tracking error and planning times.

Our planner is able to generate feasible trajectories in under 40ms, with the majority of planning times taking under 10ms, thereby enabling quick reactions to changing environments. 
All trajectories generated were feasible, and tracking errors remained under 8 degrees for the worst offending joints. 

\section{Conclusion and Future Work}
In this work, we present a vectorized manifold-constrained sampling-based motion planner that is capable of generating constraint-satisfying motion plans on the order of milliseconds, achieving over 500$\times$ speedup in challenging, cluttered, and high-dimensional problems.
We also demonstrate the scalability of our approach by tackling a real-time, whole body control problem with a 28-DoF Digit humanoid, demonstrating reactive real-time constraint-satisfying planning in dynamic environments. 
We believe that this capability opens up many possibilities for real-world tasks, and potentially for tackling more complex planning problems such as whole-body task and motion planning for humanoid robots.

For future work, we are interested in investigating extensions to the manifold-constrained formulation that could be extended to kinodynamic planning, moving beyond the limitations of quasi-static planning.
There are also improvements to the current planner that could be explored that are known in the literature, such as more sophisticated optimization techniques and integrating with continuation based methods.
We also plan to investigate dynamic constraint and robot compilation, as currently our approach requires knowledge of constraints \emph{a priori} to generate efficient SIMD kernels.

\ifanonymous
\else
\section{Acknowledgement}
We thank Thomas Cohn and Jiaming Hu for their help with baseline implementations and comparisons, and Jolie Lai for hardware experiments.
\fi

\printbibliography{}
\end{document}

%% file: preamble.tex
\usepackage[utf8]{inputenc}
\usepackage[T1]{fontenc}
\usepackage[english]{babel}
\usepackage{booktabs}
\usepackage{tabularx}
\usepackage{multirow}
\usepackage{listings}
\usepackage{subcaption}
\usepackage{xcolor}
\usepackage{xspace}
\usepackage{amsmath}
\usepackage{amssymb}
\usepackage{amsthm}

\newtheoremstyle{main}
{1em}                                                %
{1em}                                                %
{\itshape}                                           %
{0pt}                                                %
{\scshape}                                           %
{}                                                   %
{2pt}                                                %
{\thmname{#1}\thmnumber{ #2} [{\thmnote{\itshape #3}}]:~} %

\theoremstyle{main}

\usepackage{graphicx}
\usepackage{booktabs}
\usepackage{siunitx}

\usepackage{svg}
\usepackage{adjustbox}

\usepackage{csquotes}
\usepackage{algorithm}
\usepackage{algpseudocodex}

\usepackage{etoolbox}
\usepackage[
	maxbibnames=99,
	maxcitenames=2,
	natbib=true,
	style=numeric-comp,
	backend=biber,
	sorting=none,
	giveninits=true,
	url=false,
	doi=false,
	eprint=false,
	isbn=false,
]{biblatex}

\definecolor{purduegold}{HTML}{C28E0E} %

\makeatletter
\let\NAT@parse\undefined
\makeatother
\usepackage[pdfa,colorlinks,bookmarksopen,bookmarksnumbered,allcolors=purduegold]{hyperref}

\usepackage[nameinlink,capitalise]{cleveref}
\crefname{line}{line}{lines}
\Crefname{line}{Line}{Lines}
\makeatletter
\crefname{ALG@line}{line}{lines}
\Crefname{ALG@line}{Line}{Lines}

\makeatletter

\let\real@label\label

\newcommand{\alg@cref@label}[1]{%
  \begingroup
    \edef\@currentlabel{\theALG@line}%
    \real@label{#1}%
  \endgroup
}

\AtBeginEnvironment{algorithmic}{%
  \let\label\alg@cref@label
}

\makeatother
\makeatletter
\newcommand{\lineref}[2]{%
  Alg~\ref{#1}, Line~\ref{#2}%
}
\makeatother

\crefname{figure}{Fig.}{Figs.}
\Crefname{figure}{Fig.}{Figs.}
\crefname{equation}{Eq.}{Eqs.}
\Crefname{equation}{Eq.}{Eqs.}
\crefname{section}{Sec.}{Secs.}
\Crefname{section}{Sec.}{Secs.}
\crefname{definition}{Def.}{Defs.}
\Crefname{definition}{Def.}{Defs.}
\crefname{algorithm}{Alg.}{Algs.}
\Crefname{algorithm}{Alg.}{Algs.}
\crefname{assumption}{Asm.}{Asms.}
\Crefname{assumption}{Asm.}{Asms.}
\crefname{subassumption}{Asm.}{Asms.}
\Crefname{subassumption}{Asm.}{Asms.}
\Crefname{problem}{Problem}{Problems}
\crefname{problem}{Problem}{Problems}

\usepackage{flushend}

\definecolor{setgreen}{HTML}{66c2a5}
\definecolor{setorange}{HTML}{D18F00}
\definecolor{setblue}{HTML}{8da0cb}
\definecolor{setpurp}{HTML}{e78ac3}
\definecolor{setorangevariant}{HTML}{cc4400}
\definecolor{setgreenorange}{HTML}{A6B153}

\definecolor{rblue}{HTML}{4169E1}
\newcommand{\contrib}[1]{{\color{rblue} #1}}

\newcommand{\ompl}{{\color{setpurp}\textbf{OMPL}}\xspace}
\newcommand{\omplvariant}[1]{{\color{setgreenorange}\textbf{OMPL #1}}\xspace}
\newcommand{\mcvamp}{{\color{setorange}\textbf{McVAMP}}\xspace}
\newcommand{\curobo}{{\color{setgreen}\textbf{CuRobo}}\xspace}

%% file: crrtc_algo.tex
\begin{algorithm}[b]
\begin{footnotesize}
\caption{Vectorized Manifold-Constrained RRT-Connect}
\label{alg:crrtc}
\begin{algorithmic}[1]
\Require $q_{a}, q_{b}, \mathcal{F}$

\State $T_{\text{a}} \gets \text{Tree}(q_{a})$, $T_{\text{b}} \gets \text{Tree}(q_{b})$

\While{$\text{iter} < \text{MaxIterations}$}
    \State $q_{\text{rand}} \gets \text{RandomConfig}()$ \label{algline:crrtc:rand}
    \State $q^a_{\text{near}} \gets \text{Nearest}(T_{\text{a}}, q_{\text{rand}})$ \label{algline:crrtc:near}
    \State $q^a_{\text{proj}} \gets \contrib{\text{ParallelConstrainedExtend}}
    (T_{\text{a}},
    q^a_{\text{near}}, q_{\text{rand}})$ \label{algline:crrtc:extend}
    \While{$q^a_{\text{proj}} \neq \textsc{NULL}$}
        \State $q^b_{\text{near}} \gets \text{Nearest}(T_{\text{b}}, q^a_{\text{proj}})$
        \State $q^a_{\text{proj}} \gets \contrib{\text{ParallelConstrainedExtend}}(
        T_{\text{a}},
        q^a_{\text{proj}}, q^b_{\text{near}})$
        \If{ $q^a_{\text{proj}} =q^b_{\text{near}} $}
            \State $\text{P} \gets \text{ExtractPath}(T_{\text{a}},  
            T_{\text{b}})
            $
            \State \Return \textsc{P}
        \EndIf
    \EndWhile
    \State $\text{Swap}(T_{\text{a}}, T_{\text{b}})$
    \State $\text{iter} \gets \text{iter} + 1$
\EndWhile

\State \Return $\textsc{NULL}$
\end{algorithmic}
\end{footnotesize}
\end{algorithm}

%% file: parallelconstrainedextend.tex
\begin{algorithm}[!htbp]
\begin{footnotesize}
\caption{\contrib{ParallelConstrainedExtend}}
\label{alg:parallelextend}
\begin{algorithmic}[1]
\Require $T, q_{s}, q_{\text{target}}, \mathcal{F}, r, \sigma, n =$ SIMD width (e.g., 8 for Intel AVX)
\Procedure{ConstrainedExtend}{}
\State $\text{dist} \gets \min(\|q_{\text{target}} - q_s\|, r)$
\State $v_{\text{extend}} \gets \frac{q_{\text{target}} - q_s}{\|q_{\text{target}} - q_s\|}$
\State $q_{\text{steer}} \gets q_s + \text{dist} \cdot v_{\text{extend}}$ \label{algline:ext:steer}
\State $q_{\text{particles}}^{(i)} \gets q_{\text{steer}} + \epsilon_i v_{\text{extend}}, \; \epsilon_i \sim \mathcal{N}(0, \sigma^2), \; i = 1,\dots,n$ \label{algline:ext:particles}
\State $q_{\text{proj}} \gets \text{ParallelProject}(q_{\text{particles}})$ \label{algline:ext:proj}
\If{$q_{\text{proj}} = \textsc{NULL}$}
    \State \Return \textsc{NULL}
\EndIf

\If{$\neg \text{CollisionFree}(q_{\text{proj}}) \lor \|q_{\text{proj}} - q_s\| > 2 \cdot \text{dist}$} \label{algline:ext:earlyexit1}
    \State \Return \textsc{NULL}
\EndIf

\State $q_{\text{interp}} \gets n \text{ interpolated points between } q_s \text{ and } q_{\text{proj}}$ \label{algline:ext:interp}
\State $q_{\text{int\_proj}} \gets \text{ParallelProject}(q_{\text{interp}})$

\If{$\exists i: \big\|q_{\text{int\_proj}}^{(i)} - q_{\text{int\_proj}}^{(i+1)}\big\| > \frac{r}{n}$} \label{algline:ext:earlyexit2}
    \State \Return \textsc{NULL} \Comment{Projected points too far apart}
\EndIf
\LComment{Recursively interpolate and project until resolution $\delta$ is met}
\State $T.\text{addVertex}(q_{\text{proj}})$
\State $T.\text{addEdge}(q_{\text{s}}, q_{\text{proj}})$
\State \Return $q_{\text{proj}}$
\EndProcedure
\medskip
\Procedure{ParallelProject}{}
\While{\textbf{true}}
    \State $d \gets \mathcal{F}(q_{\text{particles}})$ \label{algline:ext:distance}
    \If{$\forall i: \|d^{(i)}\| < \epsilon$}
        \State \Return $q_{\text{particles}}$ \Comment{All particles converged}
    \EndIf

    \State $\text{step} \gets \text{getDescentStep}(q_{\text{particles}})$

    \If{$\exists i: \|\text{step}^{(i)}\| > \text{MaxDistance}$}
        \State \Return \textsc{NULL} \Comment{Projection diverging}
    \EndIf
    
    \State $q_{\text{particles}} \gets q_{\text{particles}} + \text{step}$
\EndWhile
\EndProcedure
\medskip
\Procedure{getDescentStep}{}
\State $q_{\text{init}} \gets q_{\text{particles}}$
\For{$\mathcal{F}_i \in \{\mathcal{F}_1, \dots, \mathcal{F}_k\}$} \label{algline:cyclic}
\State $d \gets \mathcal{F}_i(q_{\text{particles}})$
\State $J_i \gets $Jacobian of $\mathcal{F}_i(q_{\text{particles}})$
\State $\Delta q \gets \alpha J_i^T (J_i J_i^T + \lambda I)^{-1} d$ \quad \text{(or } $\alpha (J_i^T J_i + \lambda I)^{-1} J_i^T d$)
\State $q_{\text{particles}} \gets q_{\text{particles}} - \Delta q$
\EndFor

\State \Return $q_{\text{particles}} - q_{\text{init}}$

\EndProcedure

\end{algorithmic}
\end{footnotesize}
\end{algorithm}

%% file: line_plane_figs.tex
\begin{figure*}[htbp]
    \centering
        \adjustbox{trim=1cm 0.6cm 1cm 0.6cm}{
            \includeinkscape[width=\textwidth]{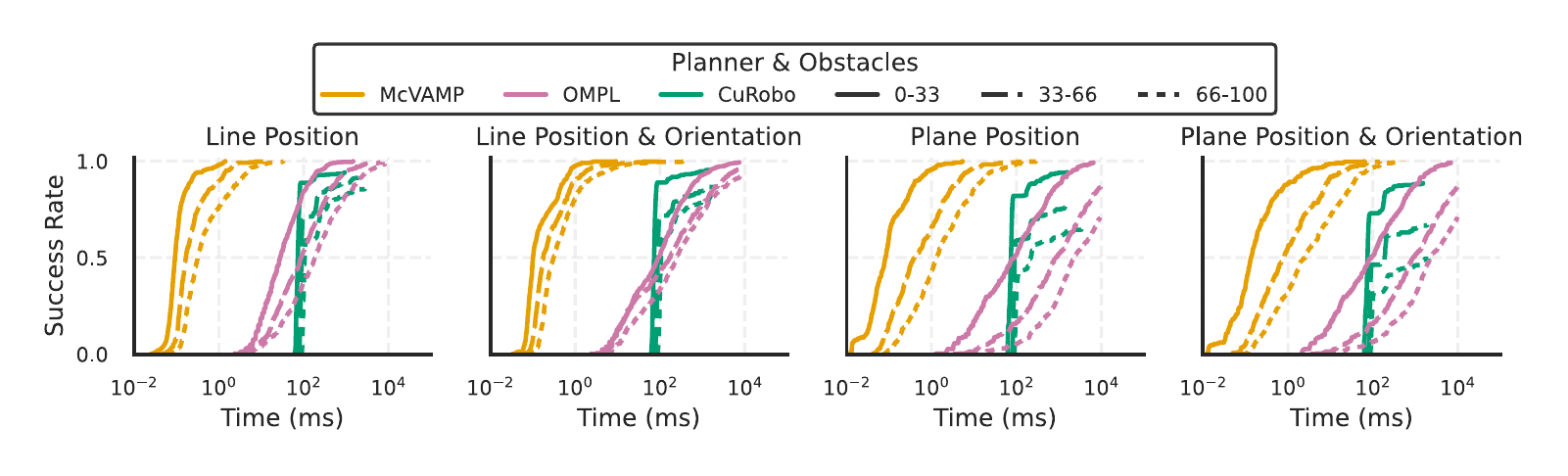}
        }
        \caption{CDF plot benchmarking planner performance for the line and plane constrained problems, binned by number of environment obstacles. \textit{1. LP - Line with Position Constraint only, 2. LPO - Line with Position and Orientation Constraint, 3. PP - Plane with Position Constraint, 4. PPO - Plane with Position and Orientation Constraint} For most problems, \mcvamp is \textbf{1000} times faster than the baselines achieving a 100\% success rates, while the other planners suffer in the success rate as the number of obstacles increases.}
        \label{fig:line_plane_plots}
        \vspace{-1em}
\end{figure*}

%% file: maze.tex
\begin{table}[h]
    \centering
    \footnotesize
    \resizebox{\columnwidth}{!}{%
        \begin{tabular}{l r rrrr r}
            \toprule
            \textbf{Method} & \textbf{Mean} & \textbf{Q1} & \textbf{Median} & \textbf{Q3} & \textbf{95\%} & \textbf{Succ.} \\
            \midrule
            \midrule
            \ompl & 25.10 & 7.33 & 17.323 & 29.20 & 85.82 & 72\% \\
            \omplvariant{(comp.)} & 13.25 & 1.00 & 3.29 & 16.80 & 52.07 & 76\% \\
            \mcvamp & \textbf{0.032} & \textbf{0.002} & \textbf{0.006} & \textbf{0.014} & \textbf{0.058} & \textbf{82\%} \\
            \bottomrule
        \end{tabular}%
    }
    \caption{Results for maze Solving. All times are reported in \textbf{seconds}. \omplvariant{(comp.)} refers to the \ompl planner with our compiled projection.}
    \label{tab:maze_table}
\end{table}

%% file: bimanual.tex
\begin{table}[h]
    \centering
    \small
    \setlength{\tabcolsep}{4pt} %
    \begin{tabular}{@{} l ccc @{}}
        \toprule
        \textbf{Method} & \textbf{T $\to$ M} & \textbf{M $\to$ B} & \textbf{B $\to$ T} \\
        \midrule
        IK-BiRRT       & 96.74 $\pm$ 48.48          & 90.11 $\pm$ 21.70          & 121.90 $\pm$ 53.61 \\
                       &  (3.69 $\pm$ 0.63) &  (5.64 $\pm$ 0.96) &  (6.26 $\pm$ 1.09) \\
        \midrule
        IK-GCS         & 63.47 $\pm$ 1.95          & 44.18 $\pm$ 2.74          & 77.87 $\pm$ 1.19 \\
                       &  (\textbf{2.52 $\pm$ 0.0}) &  (\textbf{4.73 $\pm$ 0.0}) &  (\textbf{5.82 $\pm$ 0.0}) \\
        \midrule
        \mcvamp         & 2.11 $\pm$ 1.16        & 3.53 $\pm$ 2.44        & 2.36 $\pm$ 1.63 \\
        (1-step)        &  (7.72 $\pm$ 1.50) &  (6.08 $\pm$ 1.51) &  (8.93 $\pm$ 1.33) \\
        \midrule
        \mcvamp         & \textbf{1.65 $\pm$ 0.53} & \textbf{3.49 $\pm$ 1.21} & \textbf{1.37 $\pm$ 0.65} \\
                       &  (7.62 $\pm$ 1.47) &  (7.69 $\pm$ 2.97) &  (8.28 $\pm$ 1.34) \\
        \bottomrule
        \addlinespace[1ex]
        \multicolumn{4}{l}{\textit{T: Top, B: Bottom, M: Middle.}}
    \end{tabular}
    \caption{KUKA IIWA Bimanual Planning Results: Time in \textbf{milliseconds} (top) and Path Cost (bottom in parenthesis).}
    \label{tab:bimanual}
\end{table}

%% file: humanoid_transport.tex
\begin{table}[b]
    \centering
    \scriptsize 
    
    \setlength{\tabcolsep}{2.5pt} 
    
    \begin{tabular}{l cccc}
        \toprule
        \textbf{Motion phases} & \textbf{S $\to$ T} & \textbf{T $\to$ B} & \textbf{B $\to$ M} & \textbf{M $\to$ T} \\
        \midrule
        \textbf{Tracking error (deg)} \\
        Left Knee Joint    & 1.70 $\pm$ 1.57 & 0.64 $\pm$ 0.38 & 1.28 $\pm$ 0.34 & 1.00 $\pm$ 0.50  \\
        Left Tarsus Joint  & 4.37 $\pm$ 2.26 & 4.80 $\pm$ 1.82 & 8.63 $\pm$ 1.73 & 6.38 $\pm$ 1.31  \\
        Right Knee Joint   & 3.81 $\pm$ 1.78 & 2.57 $\pm$ 1.05 & 5.61 $\pm$ 0.71 & 4.51 $\pm$ 0.63  \\
        Right Tarsus Joint & 3.51 $\pm$ 1.73 & 5.15 $\pm$ 1.93 & 7.74 $\pm$ 1.93 & 5.85 $\pm$ 1.89  \\
        \midrule
        \textbf{Planning Time (ms)} & 4.93 $\pm$ 0.12 & 10.11 $\pm$ 1.65 & 13.42 $\pm$ 3.68 & 16.28 $\pm$ 2.11 \\
        \bottomrule
        \addlinespace[1ex]
        \multicolumn{5}{l}{\textit{Note: S: Standing Pose, T: Top, B: Bottom, M: Middle.}}
    \end{tabular}
    \caption{Planning to transport a box. The robot picks up the box from the top shelf, transports it to the bottom shelf, then the middle and finally the top one. We show the mean tracking error in the legs between the controller and the planner, and planning time in \textbf{milliseconds}. All numbers are reported as mean $\pm$ 1 standard deviation.}
    \label{tab:humanoid_shelf_planning}
\end{table}